\documentclass[letterpaper, 10 pt, conference]{ieeeconf}  %

\IEEEoverridecommandlockouts                              %
\usepackage{times}

\usepackage{multicol}
\usepackage[bookmarks=true]{hyperref}
\usepackage{graphicx}
\usepackage{subcaption}

\usepackage{amsmath}
\usepackage{amssymb}
\usepackage{bm}

\usepackage{siunitx}

\usepackage[
backend=biber, 
bibencoding=utf8,
style=ieee, 
sorting=none, 
natbib=true, 
doi=false, 
isbn=false, 
url=false, 
eprint=false, 
maxcitenames=1, 
mincitenames=1,
dashed=false
]{biblatex}

\usepackage{xspace}
\newcommand{\ie}{i.e.,\xspace}
\newcommand{\eg}{e.g.,\xspace}

\usepackage[printonlyused]{acronym}
\acrodef{ICP}{Iterative Closest Point}
\acrodef{GNSS}{Global Navigation Satellite System}
\acrodef{HMI}{Hazardously Misleading Information}

\usepackage{lipsum}
\usepackage[dvipsnames]{xcolor}

\newcommand{\R}{\mathbb{R}}
\newcommand{\bx}{\bm{x}}

\newcommand{\hbx}{\hat{\bm{x}}}
\newcommand{\by}{\bm{y}}
\newcommand{\Bf}{\bm{f}}

\newcommand{\bw}{\bm{w}}

\newcommand{\be}{\bm{e}}
\newcommand{\bp}{\bm{p}}
\newcommand{\bq}{\bm{q}}
\newcommand{\bn}{\bm{n}}
\newcommand{\bt}{\bm{t}}
\newcommand{\bphi}{\bm{\phi}}
\newcommand{\bA}{\bm{A}}
\newcommand{\bI}{\bm{I}}

\newcommand{\bQ}{\bm{Q}}
\newcommand{\bD}{\bm{g}^T}
\newcommand{\bH}{\bm{H}}

\newcommand{\bR}{\bm{R}}
\newcommand{\bSigma}{\bm{\Sigma}}
\newcommand{\bSigmaAlpha}{\bm{\Sigma}_\alpha}

\renewcommand{\P}[1]{p\left(#1\right)}
\newcommand{\norm}[1]{\left\lVert#1\right\rVert}

\DeclareMathOperator*{\argmin}{arg\,min}
\DeclareMathOperator{\sign}{sgn}

\usepackage{tabularx}
\usepackage{booktabs}

\AtBeginDocument{}%
\AtBeginDocument{}%

\DeclareSourcemap{
  \maps[datatype = bibtex]{
    \map{
       \step[fieldsource = abstract,
          match = \regexp{([^\\])\%},
          replace = X]
    }
  }
}

\title{\LARGE \bf
  Toward Certifying Maps for Safe Registration-based Localization Under Adverse Conditions
}

\author{Johann Laconte$^{1}$, Daniil Lisus$^{1}$ and Timothy D. Barfoot$^{1}$%
\thanks{$^{1}$ University of Toronto Institute for Aerospace Studies (UTIAS), 4925 Dufferin St, Ontario, Canada.
        {\tt\small \{johann.laconte, daniil.lisus\}@robotics.utias.utoronto.ca, 
         tim.barfoot@utoronto.ca}%
}}

\begin{document}

\maketitle
\thispagestyle{empty}
\pagestyle{empty}

\begin{abstract}
    In this paper, we propose a way to model the resilience of
    the \ac{ICP} algorithm in the presence of
    corrupted measurements.
    In the context of autonomous vehicles, certifying the safety of the localization process poses a significant challenge.
    As robots evolve in a complex world, various types of noise can impact the measurements.
    Conventionally, this noise has been assumed to be distributed according to a zero-mean Gaussian distribution. 
    However, this assumption does not hold in numerous scenarios, including adverse weather conditions, occlusions caused by dynamic obstacles, or long-term changes in the map. 
    In these cases, the measurements are instead affected by large and deterministic faults.
	This paper introduces a closed-form formula approximating the pose error resulting from an ICP algorithm when subjected to the most detrimental adverse measurements.
    Using this formula, we develop a metric to certify and pinpoint specific regions within the environment where the robot is more vulnerable to localization failures in the presence of faults in the measurements.
\end{abstract}

\section{Introduction}
  \label{sec:intro}
  Reliable localization is a vital task for self-driving robots, as it plays a key role in ensuring their safety and effective operation.
However, sensors are susceptible to making errors, necessitating the implementation of robust safeguards to mitigate potential risks.
In this paper, we focus on the domain of range-based localization, a technique relying on range sensors such as lidar or radar.
In addition to being inherently noisy, these sensors can also encounter large, deterministic errors that heavily deviate from the common Gaussian noise assumption, posing significant challenges to accurate localization.

Measurement corruptions can be caused by diverse sources.
For example, an occlusion may cause a portion of the environment to be completely blocked from the sensor's view, leading to missing or inaccurate range readings.
Similarly, adverse weather conditions, such as heavy snowstorms or fog, can distort the sensor measurements in a nonrandom manner \cite{Courcelle2022}.
Consistent measurement errors can also arise from long-term changes in an outdated map, including the presence of new buildings or parked cars.
These factors result in systematic and deterministic errors that are significantly different from the statistical properties of Gaussian noise.

\begin{figure}[t!]
  \includegraphics[width=\linewidth]{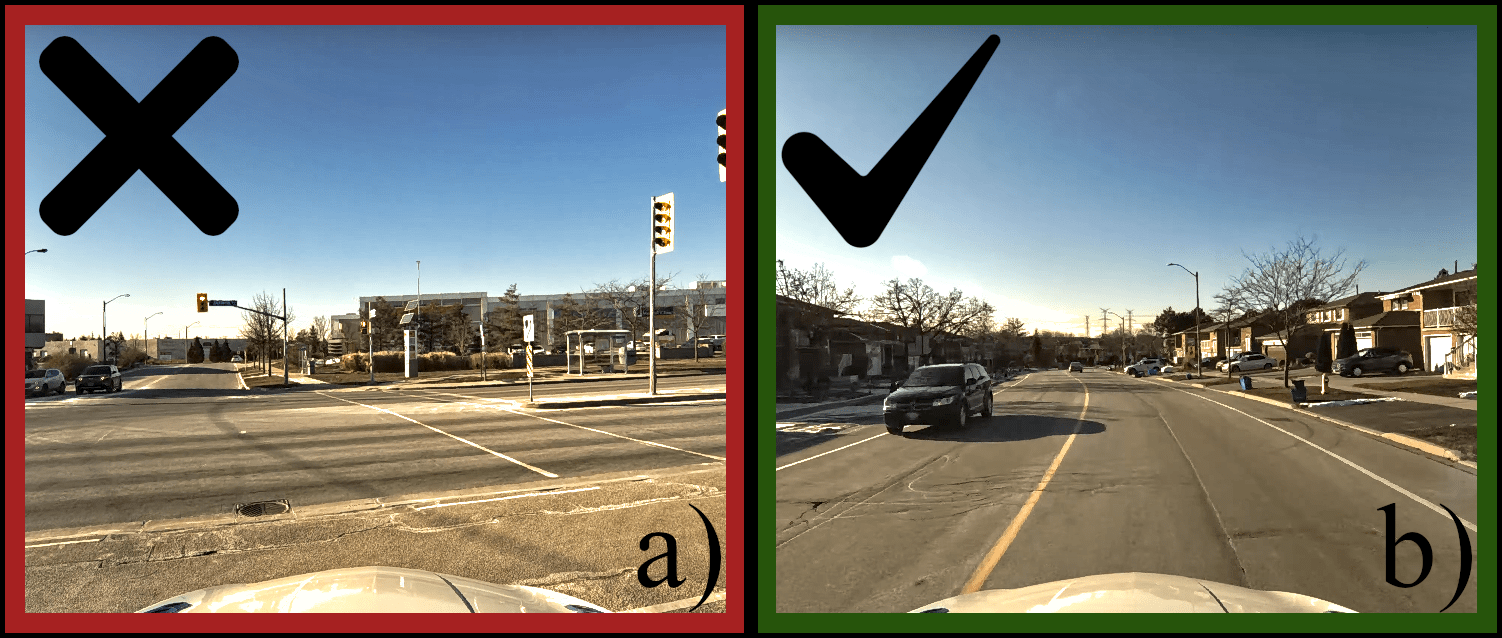}
  \caption{
    In the context of pose estimation, specific environments entail higher risks compared to others. 
    For instance, a large intersection a) offers limited landmarks for localization, whereas a suburban area b) presents numerous houses and landmarks.
    Consequently, occlusions or map alterations at the intersection may lead to a significantly larger pose estimation error compared to the suburban area.
  }
  \label{fig:intro}
  \vskip-1em
\end{figure}

The \ac{ICP} algorithm is established as one of the most popular approaches for estimating a robot's pose using range sensors \cite{Pomerleau2015}.
Depending on the map against which \ac{ICP} is trying to localize, a corruption of the pointcloud can be more or less impactful.
\autoref{fig:intro} depicts some examples from the self-driving application.
In a large intersection, few structures are close enough to provide good constraints for \ac{ICP}.
As such, any faulty measurement on the structures would have a great impact on the localization output.
On the contrary, smaller streets with lots of houses and other obstacles provide a much safer environment, as \ac{ICP} can now rely on a variety of landmarks to localize.
Even though \ac{ICP} is frequently equipped with robust outlier rejection features, some faults can still be considered as inliers and impact the accuracy of the localization process.
While not necessarily large enough to break the localization process, these inlier faults can degrade the localization accuracy to a degree that may easily result in a crash in autonomous driving scenarios.
Our paper aims to quantify the map-dependent resilience of the \ac{ICP} localization algorithm.
Our contributions are 
1) a closed-form formula that approximates the worst possible error on the pose for a given amount of corrupted points;
2) a visualization of the worst corruption by applying the faults to the measured pointcloud; and
3) a quantification of the resilience of \ac{ICP} for a specific map against corrupted measurements. 
We provide a quantitative analysis of our framework, as well as a qualitative evaluation of both structured and unstructured environments, showing that our framework can pinpoint dangerous locations in the event of corrupted measurements.

\section{Related Work}
  \label{sec:rw}
Extensive research has been conducted in the field of aerospace engineering to address safety considerations.
In this context, GNSS measurements are primarily used for position estimation.
The safety of this estimation is based on the \ac{HMI} metric, which looks at the probability that the position estimate is sufficiently erroneous to be considered hazardous, while no fault detectors have been triggered.
The detectors can take on various forms, but two are most prevalent: residuals and solution-separation methods \cite{Joerger2014, Oliveira2022}.
The residuals-based methods look at the measurement residuals coming from the estimation algorithm \cite{Zhao2021}, such as the innovation in a Kalman filter \cite{Hage2022}.
When such residuals are higher than a threshold, a fault in the measurements is likely and an alarm is triggered.
On the other hand, solution-separation-based methods try to isolate and discard any potentially faulty measurement.
Recently, \citet{Arana2019} proposed to adapt these methods to robotics.
They used the \ac{HMI} framework to monitor the safety of a landmark-based localization pipeline. Their work was adapted to batch estimation by \citet{Hafez2020}.
Finally, \citet{Chen2020} used this framework to propose a way to enhance the safety of maps by adding well-placed landmarks in the environment. 
All of these previous efforts aim at certifying that the system will be able to detect faults, which is fundamentally different from certifying the well-being of the system.
Indeed, a very poor map can be classified as safe as long as is it easy for the system to detect that its estimated pose is not right.
As such, we propose a novel method to directly monitor the reliability of the localization process, and not its capability to detect hazardous events.

In this context, the notion of \textit{localizability} is the standard for range-based localization.
Localizability is defined as the capability of the localization system to produce a good estimate of the robot's pose.
This information is paramount in under-constrained environments, such as tunnel-like surroundings \cite{Tuna2022}, where the localization process is prone to yielding bad estimates.
\citet{Nubert2022} proposed a way to learn the localizability in underground environments.
\citet{Ebadi2021} developed a method to monitor the degeneracy of \ac{ICP} throughout the localization process by looking at the condition of the measurement matrix after linearizing the system.
Similarly, \citet{Zhang2019} proposed to model the safety of the localization in an environment using Fisher information.
Finally, \citet{Aldera2019} proposed to train a classifier to automatically label odometry estimates as good or bad, showing that rejecting poor solutions leads to a better overall estimate.
\citet{Carson2022} used a similar idea to classify the integrity in visual localization, removing inaccurate estimates.
However, the localizability analysis does not encompass the possibility of corrupted measurements that can arise in numerous situations, such as adverse weather or occlusion from other dynamic obstacles.
Indeed, the analysis relies solely on the measured pointcloud, without taking into account the possibility that large errors can add malicious information.
We propose a way to not only examine the safety of the system under nominal conditions, but also its performance in situations where certain measurements may provide adversarial information to the robot.

Within the area of compromised data, different papers explicitly assess the resistance of algorithms to corruption by generating, altering, or removing some measurements.
\citet{Kong2023} designed a benchmark that directly takes into account the corruption of data from diverse sources, such as the weather or sensor failures.
\citet{Xiang2018} proposed a method to generate and perturb pointclouds so as to lead a classification network to mislabel scanned objects.
\citet{Cao2019} developed a method to craft 3D obstacles that were undetectable by deep learning algorithms working with lidar data.
\citet{Delecki2022} designed a method to craft disturbances in lidar readings and characterize the failures of the perception systems in adverse weather conditions.
To the best of our knowledge, \citet{Yoshida2022} is the only approach that directly attacks a pointcloud to force the \ac{ICP} algorithm to move to a desired target position instead of the ground truth.
However, the choice of the desired target is left to the user, and thus no guarantee can be automatically inferred from the framework.
As such, we propose a closed-form algebraic approach to automatically identify the most unfavorable pose achievable through \ac{ICP} via corruption of the measurements.

\section{Preliminaries}
  \label{sec:theory}
First, we introduce the methodology for modeling corruption in conventional estimation problems, then propose a safety metric on these systems.
Finally, we offer a brief illustration of the process to transform pointcloud alignment into an approximated linear problem.

\subsection{Problem Statement}\label{sec:prob_ref}
In robotics, many problems can be formed as a least-squares minimization.
In particular, we are interested in standard linear measurement problems of the form
\begin{equation}\label{eq:standard_prob_form}
  \begin{aligned}
    \by = \bA\bx + \bw,
  \end{aligned}
\end{equation}
where $\by\in\R^m$ is the measurement, $\bx\in\R^n$ is the state to estimate, $\bA\in\R^{m\times n}$ is the matrix linking the state to the measurement, and $\bw\in\R^m$ is random Gaussian noise with covariance $\bSigma$.
This formulation assumes that the error on the measurement can be modelled as a zero-mean, Gaussian random variable.
Perturbations resulting from inclement weather or occlusions from other dynamic obstacles can contradict this assumption.
Therefore, as is done in \cite{Hafez2020}, we assume a measurement can be subject to both noise and deterministic, possibly large, faults (corruptions):
\begin{equation}\label{eq:full_meas_model}
  \by = \bA\bx + \bw + \bQ\Bf,
\end{equation}
where $\Bf\in\R^{n_f}$ are the faults, and $\bQ\in\R^{m\times n_f}$ is a sparse matrix of zeros and ones applying the faults to the associated measurements.
As such, a part of the measurements is subject to possibly large faults that can hinder the estimation.

In the context of linear estimation, the solution to a least-squares problem can be written as %
\begin{equation}
\begin{aligned}
  \hat{\bx} &= \argmin_{\bx} \sum_{i=1}^m \alpha_i\cdot(\by_i - \bA_i\bx)^T\bSigma_i^{-1}(\by_i - \bA_i\bx) \\
  & \quad =\argmin_{\bx} (\by - \bA\bx)^T\bSigmaAlpha^{-1}(\by - \bA\bx),
\end{aligned}
  \label{eq:linear_prob}
\end{equation}
where $\by_i$ is the $i^\text{th}$ measurement with component $\bA_i$ and covariance $\bSigma_i$, and $\bx$ is the pose to estimate.
The measurement $\by_i$ can be either a scalar or a multidimensional quantity.
Note that each measurement is weighted by a factor $\alpha_i$, which is used to control the impact of each measurement on the estimate and is incorporated into the lifted matrix $\bSigmaAlpha$.

In the case of systems of the form defined in \eqref{eq:linear_prob}, the solution is given by
\begin{equation}
  \hbx = \bH\by, \quad\text{with }\bH = \left(\bA^T\bSigmaAlpha^{-1}\bA\right)^{-1}\bA^T\bSigmaAlpha^{-1}.
\end{equation}
The corresponding state-estimation error can be computed as
\begin{equation}
  \begin{aligned}
    \be &= \hbx - \bx\\
        &= \bH\left(\bA\bx+\bw+\bQ\Bf\right) - \bx \\
        &= \bH\left(\bw + \bQ\Bf\right).
  \end{aligned}
  \label{eq:error_linsys}
\end{equation}
The faults $\bQ\Bf$, unlike the noise $\bw$, are not zero-mean random variables, but are unknown, possibly large, deterministic quantities.
As such, the faults generate a bias in the estimation that could be harmful.

Finally, robust filters have the task of discriminating between inlier and outlier measurements.
Among them, the trimmed distance filter is one of the most popular \cite{Pomerleau2015}.
Given an initial guess $\bx_0$, the trimmed distance filter removes any measurement that has a corresponding error that is larger than a fixed distance $d$:
\begin{equation}
  \alpha_i = 
  \begin{cases}
    1 &\text{if } \norm{\by_i - \bA_i\bx_0}_\infty \leq d, \\
    0 & \text{otherwise}.
  \end{cases}
  \label{eq:trimming_weights}
\end{equation}
As such, any fault trying to corrupt a system making use of a trimmed distance filter has to be small enough to be deemed an inlier.
Here it is assumed that the trim distance is tuned correctly, in that it does not remove useful points and only serves to guard against outliers. 

\subsection{Safety Metric}\label{sec:safety_metric}
Aiming at certifying a localization algorithm, we define a safety metric as the error on the pose being below a certain threshold:
\begin{equation}
  |e_j| = |\bD_j(\hbx - \bx)| \leq r_j, %
  \label{eq:error}
\end{equation}
where $\hbx$ is the estimated state, $\bx$ is the ground truth, $\bD_j$ is a row matrix that extracts the $j^\text{th}$ coordinate from the state, and $r_j$ is the error-specific threshold.
As such, an estimate is considered safe if each of its components has an error below a certain threshold $r_j$.
Note that we choose to certify each component of the estimated state independently instead of looking at the $L_2$ norm of the error.
This is particularly useful in self-driving applications where, for instance, the lateral error is often more important than the longitudinal one.
Using this definition, we seek to compute the probability that a pose estimate is safe:
\begin{equation}
  \P{|e_j| \leq r_j} \geq p_\text{safe}.
  \label{eq:cond_safe}
\end{equation}
In other words, a given estimate is said to be certified if its probability to be in the safe zone of radius $r_j$ is above $p_\text{safe}$.
In the presence of faults, we are interested in the maximum number of faulted measurements that can happen at the same time before \eqref{eq:cond_safe} becomes false.
This quantity is defined as the \textit{resilience} of the system.

\subsection{ICP Formulation}
Sections \ref{sec:prob_ref}, \ref{sec:safety_metric}, and all of Section \ref{sec:corr_meas}, are presented generally for problems of form \eqref{eq:full_meas_model}.
To verify and show the usability of the theory, we select the popular class of autonomous driving localization approaches based on the ICP algorithm as an example.
  First, we briefly show a linear simplification of ICP.
  For one iteration, the measurement model is written as
  \begin{equation}
    \begin{aligned}
      \bq_i &= \bR(\bp_i + \bw_i) + \bt,
    \end{aligned}
    \label{eq:ICP_lin_1}
  \end{equation}
  where $\bp_i$ and $\bq_i$ are the points in the sensor and map frames respectively, and $\bw_i\sim\mathcal{N}(\bm{0}, \sigma^2\bm{I})$ is the noise on the measured point $\bp_i$, $\bm{I}$ being the identity matrix.
  The matrix $\bR$ and vector $\bt$, respectively, denote the rotation and translation components of the pose.
  Assuming the rotation $\bR$ between the scan and map pointclouds is small, we use the small-angle approximation $\bR\approx\bm{I}+\bphi^\wedge$ and linearize \eqref{eq:ICP_lin_1} as
  \begin{equation}
    \begin{aligned}
            \bq_i &\approx (\bm{I} + \bphi^\wedge)\bp_i + \bR\bw_i + \bt\\
            &= \bp_i - \bp_i^\wedge\bphi + \bR\bw_i + \bt  \\
            &= \begin{bmatrix}
                  \bm{I} & -\bp_i^\wedge 
                \end{bmatrix}
                \begin{bmatrix}
                  \bt \\ \bphi
                \end{bmatrix}
                + \bp_i + \bw_i',
    \end{aligned}
    \label{eq:ICP_lin}
  \end{equation}
  where $(\cdot)^\wedge$ is a cross-product operator transforming the vector to a 3$\times$3 skew-symmetric matrix \cite{Barfoot2017}.
  In the case of point-to-plane \ac{ICP} with unit normals $\bn_i$ in the map frame, we project the measured points $\bp_i$ onto the associated normals $\bn_i$.
  Reworking \eqref{eq:ICP_lin} and stacking the $N$ measurements into one vector $\by$, we have the linear system
  \begin{equation}
    \underbrace{\begin{bmatrix}
        \bn_1^T(\bq_1-\bp_1) \\
        \vdots \\
        \bn_N^T(\bq_N-\bp_N)
    \end{bmatrix}}_{\by} =%
                \underbrace{\begin{bmatrix}
                    \bn_1^T & -\bn_1^T\bp_1^\wedge\\
                  \vdots & \vdots \\
                  \bn_N^T & -\bn_N^T\bp_N^\wedge\\
              \end{bmatrix}}_{\bA}
                \underbrace{\begin{bmatrix}
                  \bt \\ \bphi
              \end{bmatrix}}_{\bx}
              + \,\bw, %
              \label{eq:ICP_ptpl}
  \end{equation}
  of form \eqref{eq:standard_prob_form} with $\bw\sim\mathcal{N}(\bm{0}, \sigma^2\bm{I})$.
  This formulation has two main assumptions: known data association and a noniterative approach.
  While these assumptions may appear restrictive, we demonstrate in \autoref{subsec:approx} that they establish a suitably conservative approximation of the error of the real, nonapproximated \ac{ICP} algorithm.

\section{Corrupted Measurements}
In this section, we provide a closed-form formula for the worst pose estimate given corrupted measurements from a defined set.
This closed form is used to compute the probability that a pose estimate is hazardous, as defined in \eqref{eq:cond_safe}.
Then, we define a metric to certify the resilience of \ac{ICP} in different scenarios.
\label{sec:corr_meas}
\subsection{Worst Pose Estimate}%
  This section proposes a closed-form formula to compute the probability of a pose estimate being hazardous, given corrupted measurements from a defined set.
  First, we define the constraint that the faults need to satisfy so that they are not trimmed by the outlier filter.
  As defined in \eqref{eq:full_meas_model}, the measurement $\by$ contains both a probabilistic zero-mean noise $\bw$ and deterministic faults from a defined set $\bQ\Bf$.
  Faults that maximize the pose error will be such that they are considered inliers according to \eqref{eq:trimming_weights}, but are otherwise maximally detrimental to the accuracy of the pose estimate.
  It is assumed that the initial guess is close enough to the ground truth so that the pose error is primarily dominated by noise and inlier faults.
  Relaxing this assumption is left for future work.
Using \eqref{eq:full_meas_model} and \eqref{eq:trimming_weights} we have the following constraint on the faults:
  \begin{equation}
    \begin{aligned}
      &\norm{\by - \bA\bx_0}_\infty \leq d \\
      \Leftrightarrow&\norm{\bw + \bQ\Bf}_\infty \leq d,
    \end{aligned}
    \label{eq:constraint_meas}
  \end{equation}
  meaning that all errors on the measurements are below the outlier detection threshold $d$.
  However, this constraint acts on the whole measurement vector and not only on the corrupted subset.
  As such, assuming the noise is small enough not to trigger the outlier detector, we reduce the constraint to only act on the subset of faulted measurements, as
  \begin{equation}
    \norm{\bQ^T\bw + \Bf}_\infty \leq d,
    \label{eq:constraint}
  \end{equation}
  where $\bQ^T\in\R^{n_f\!\times m}$ extracts the noise components of the faulted measurements from the noise vector $\bw\in\R^m$.
  As such, the constraint \eqref{eq:constraint} forces the faults to be small enough that the faulted measurements are still considered inliers.
  Using this constraint, we now seek to find the worst pose error that a defined set of faulted measurements could produce.
  Following \eqref{eq:error_linsys} and \eqref{eq:error}, the maximum error that a set of faulted measurements can induce on the pose while being undetected by the outlier detector is defined by
  \begin{equation}
    \begin{aligned}
      &\max_{\norm{\bQ^T\bw + \Bf}_\infty \leq d} |e_j| = \max_{\norm{\bQ^T\bw+\Bf}_\infty \leq d} |\bD_j\bH\left(\bw + \bQ\Bf\right)| \\[.4em]
               &=\max_{\norm{\bQ^T\bw+\Bf}_\infty \leq d}  |\bD_j\bH\bar{\bQ}\bw + \bD_j\bH\bQ\left(\bQ^T\bw + \Bf\right)|, %
    \end{aligned}
    \label{eq:main_derivation2}
  \end{equation}
  where $\bar{\bQ}=\bI - \bQ\bQ^T$.
  In order to find a closed form, we recall that for all $a\in\R$ and $x\in[-m,m]$ we have
  \begin{equation}
    \max_{x\in[-m,m]} |a+x| =  \max \left\{ |a\pm m| \right\}.
    \label{lemma:convex}
  \end{equation}
  As such, using \eqref{lemma:convex}, the worst error on the pose \eqref{eq:main_derivation2} can be rewritten as
  \begin{equation}
    \max_{\norm{\bQ^T\bw + \Bf}_\infty \leq d} |e_j| = \max\{\left|\bD_j\bH\bar{\bQ}\bw \pm m\right|\},
    \label{eq:main_derivation3}
  \end{equation}
  with 
  \begin{equation}
  \begin{aligned}
    m &= \max_{\norm{\bQ^T\bw+\Bf}_\infty \leq d}\bD_j\bH\bQ\left(\bQ^T\bw + \Bf\right) \\
      &= d\norm{\bD_j\bH\bQ}_\infty,
  \end{aligned}
  \label{eq:def_m}
  \end{equation}
  where we used the fact that for any vectors $\bm{a}, \bm{b}$, we have
  \begin{equation}
    \max_{\norm{\bf{b}}_\infty \leq 1} \bm{a}^T\bm{b} = \sum_k \left|a_{k}\right| =  \norm{\bm{a}^T}_\infty,
    \label{eq:linf_norm}
  \end{equation}
  with $\bm{a} = \begin{bmatrix}a_1 \cdots a_N\end{bmatrix}^T$.
  Plugging this back into \eqref{eq:main_derivation3}, we finally find that the worst error on the pose is defined by
  \begin{equation}
    \label{eq:error_faults}
    \begin{aligned}[b]
      \max_{\norm{\bQ^T\bw + \Bf}_\infty \leq d} |e_j| &= \max \left\{\left|\bD_j\bH\bar{\bQ}\bw \pm d\norm{\bD_j\bH\bQ}_\infty\right|\right\} \\
            &= \bD_j\bH\bar{\bQ}\bw + s \cdot d\norm{\bD_j\bH\bQ}_\infty,
    \end{aligned}
  \end{equation}
  with $s = \sign{\left(\bD_j\bH\bar{\bQ}\bw\right)}$.
  Therefore, for a given noise and set of corrupted measurements, the maximum error on the pose can be found in closed-form.
  Note, this equation yields the maximum error on the pose and not the fault vector that induces it. 

  In order to find the probability distribution of the worst error $|e_j|$, we use the notation $v=\bD_j\bH\bar\bQ\bw$ and do the following manipulation:
  \begin{equation}
    \begin{aligned}
      \P{|e_j|>r_j} &= \P{e_j>r_j , v\geq 0\right) + p\left(e_j<-r_j , v< 0}\\
                   &= 2\P{e_j>r_j , v\geq 0},
    \end{aligned}
  \end{equation}
  where we group the $v\geq0$ and $v<0$ cases as $v$ is a linear function of the noise $\bw$ and thus a zero-mean Gaussian random variable.
  Expanding the error $|e_j|$, we find
  \begin{equation}
    \begin{aligned}
      \P{|e_j|>r_j} &= 2\,\P{v+d\norm{\bD_j\bH\bQ}_\infty > r_j, v\geq0} \\
                              &= 2\,\P{v> \max\left\{r_j -d\norm{\bD_j\bH\bQ}_\infty, 0\right\}} \\
                             &= \min\left\{2\left(1 - \Phi_{\mu,\sigma}(r_j)\right), 1\right\},
    \end{aligned}
    \label{eq:proba_hazardous}
  \end{equation}
  where $\Phi_{\mu,\sigma}(\cdot)$ is the standard cumulative distribution function of the normal distribution, with 
  \begin{equation}
    \begin{aligned}
      \mu &= d\norm{\bD_j\bH\bQ}_\infty \\
      \sigma^2 &= \bD_j\bH\bar{\bQ}\bSigma\left(\bD_j\bH\bar{\bQ}\right)^T.
    \end{aligned}
  \end{equation}
  In conclusion, this section provides a way to compute efficiently the probability of having a hazardous estimate of the pose given a malicious set of corrupted measurements, using \eqref{eq:proba_hazardous}.
  We show in the next section how to certify a map using this metric, before demonstrating how the faults on the point-to-plane measurements can be linked back to the measured lidar points.

\subsection{Map Certification}
In the preceding section, we demonstrated the ability to determine efficiently the maximum localization error for a specific set of corrupted measurements.
Nevertheless, due to the large number of points in a lidar scan, exhaustively testing every combination of corrupted points becomes impractical.
To address this challenge, we propose to model the lidar field of view as a collection of 3D angular sectors splitting the $360\si{\degree}$ $x$-$y$ plane and extending infinitely into $z$.
All points within each sector are then either corrupted or unaltered. %
Although this assumption may appear restrictive, numerous events are inherently tied to a sector of corrupted points.
For instance, cars parked on the side of the street will result in a coherent shift of the measured pointcloud compared to a map that was constructed without them present.
Occlusions can also lead to a sector-wide pointcloud alteration, thus appearing as fully faulted sectors.

Using this modeling approach, we define the resilience $R$ at a specific position on the map as the maximum proportion of the lidar scan that can be corrupted before the validity of the certificate stated in \eqref{eq:cond_safe} is compromised.

\section{Experiments}
  \label{sec:experiments}
  First, we show how to link the faults back to a meaningful perturbation in the measured pointcloud. Then, we evaluate the quality of our assumptions in the context of a real, iterative \ac{ICP}.
Finally, we show the applicability of our method on real-world environments.%

\subsection{Fault Visualization}
  Although computing the explicit faults is not required for the safety assessment, it can be useful for visualization and evaluation purposes.
  As the faults are applied on the measurement equation defined in \eqref{eq:full_meas_model}, the corruption is not linked to the measured points but rather the projected distance between the lidar points and the map.
  As such, in this section, we show how to recover a corruption on the pointcloud from a set of faults originally applied to the point-to-plane measurements.
  First, we recall that the vector that maximizes \eqref{eq:linf_norm} is given by
   \begin{equation}
     \bm{b} = \sign{\bm{a}},
   \end{equation}
   where the sign function is extended to a vector, applying the sign function component-wise.
   As such, from \eqref{eq:def_m} and \eqref{eq:error_faults}, we can deduce that the fault vector $\Bf$ that maximizes the error on the pose is
   \begin{equation}
     \Bf = s\cdot d \cdot \sign{\left(\bD_j\bH\bQ\right)^T} - \bQ^T\bw.
     \label{eq:fault_vector}
   \end{equation}
  Using \eqref{eq:fault_vector}, we are able to construct a fault vector on the point-to-plane measurements.
  However, this fault vector does not directly corrupt the measured points, which are 3D quantities, but rather the projected distances.
  To model point-wise corruption, we seek to find a perturbation $\Delta \bp_k$ for each measured point $\bp_k$ subject to corruption, with a perturbed point $\bp'_k$ defined as
   \begin{equation}
     \bp'_k = \bp_k + \Delta\bp_k.
   \end{equation}
   Rewriting \eqref{eq:ICP_ptpl} for one corrupted point, we have
   \begin{equation}
     \begin{aligned}
       & \bn_k^T(\bq_k - \bp'_k) = -\bn_k^T{\bp'_k}^\wedge\bphi + \bn_k^T\bt + \bw_k \\
       \Leftrightarrow\hskip1em &\bn_k^T(\bq_k - \bp_k) = -\bn_k^T\bp_k^\wedge\bphi + \bn_k^T\bt + \bw_k \\ &\hskip8em+ \bn_k^T\Delta\bp_k - \bn_k^T\Delta\bp_k^\wedge\bphi.
     \end{aligned}
   \end{equation}
  Therefore, the fault on the point-to-plane measurement $f_k$ is linked to the corrupted point by the relation
  \begin{equation}
     \begin{aligned}
       f_k = \bn_k^T\Delta\bp_k - \bn_k^T\Delta\bp_k^\wedge\bphi, \\
     \end{aligned}
      \label{eq:rel_f_dp}
  \end{equation}
  from which we extract a viable solution
  \begin{equation}
    \Delta\bp_k = f_k\bn_k,
    \label{eq:dbp_f}
  \end{equation}
  since $\bn_k^T\bn_k^\wedge=\bm{0}$.
  Note that \eqref{eq:rel_f_dp} possesses many solutions as we retrieve a 3D quantity from a 1D fault.
  From the perspective of our framework, all solutions are equally good, as all possible corruptions lead to the same corrupted point-to-plane measurements.
  However, \eqref{eq:dbp_f} is particularly interesting as it is independent of the state variable $\bphi$.
  As such, a fault on the point-to-plane measurement can be seen as a shift of the measured point along the associated map's normal.

  \begin{figure}[t]
    \centering
    \includegraphics[width=\linewidth]{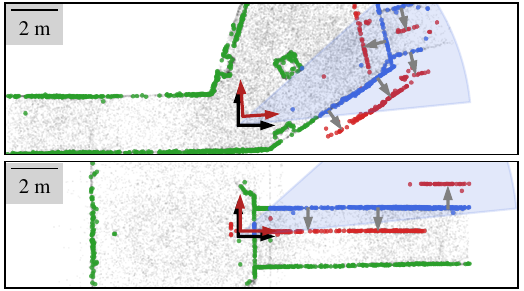}
    \caption{Examples of faulted pointclouds from underground passages of the DARPA Subterranean challenge finals \cite{ebadi2022present}. 
    The faults try to corrupt the $y$ component of the pose, with the corrupted pose estimated by ICP in red, and the ground truth pose in black.
    Light grey points correspond to the reference map.
    The measured lidar point cloud is illustrated in green, with a sector (shaded blue) being corrupted.
    The blue and red points correspond to the sector before and after corruption, with gray arrows showing the shift induced by the corruption.
    The corruption offsets points in different directions and at different angles to achieve a maximum error in the $y$ component.
    Measured points corresponding to the ceiling and floor are omitted for better visualization.
  }
    \label{fig:example_faults}
    \vspace{-0.5cm}
  \end{figure}

   As examples, \autoref{fig:example_faults} depicts two submaps of the DARPA Subterranean challenge finals \cite{ebadi2022present}, where a sector of the lidar scan is corrupted in a way to maximize the error in the $y$ component of the pose. 
   These maps depict underground environments with clear features that \ac{ICP} is using to localize.
   The inlier threshold distance has been set to $d=\SI{1}{\m}$ for visualization purposes.
   Using \eqref{eq:fault_vector}, we compute the worst faults on the $y$ component for the given corrupted sector.
   The faults are then transformed back into a corruption on the pointcloud via \eqref{eq:dbp_f}.
   In the first example (top), the corrupted points shift and slightly rotate the pose estimate to achieve the worst $y$ component estimate while evading an outlier detector.
   In the second example (bottom), the corruption takes a slightly more complex appearance.
   One might intuitively think that the worst pose estimate would come from shifting all the points in the same direction. However, our framework finds a more complex modification of the pointcloud that results in an even worse estimate.
   Indeed, a coherent shift of all points in a translation-only alignment problem would result in the worst estimate. 
   However, as \ac{ICP} has to estimate both the translation and the rotation components of the pose, shifting the whole wall in the same direction would make the \ac{ICP} solution rotate instead of translate. %
   Accordingly, the worst corruption modifies the pointcloud in such a way as to prevent rotation, resulting in a pure shift along the $y$ axis.

\subsection{Quality of Approximations}
\label{subsec:approx}
As is done in \ac{HMI}-based safety analysis (\eg \cite{Arana2019, Hafez2020}), our method reduces \ac{ICP} to a linear, noniterative problem.
This framework assumes a known data association between the map and the scan, which is impossible to have in practice.
We show in this section that our framework yields an approximate upper-bound to a real iterative \ac{ICP} algorithm, in which data association must be solved at each iteration.
To do this, we sample \SI{250}{} submaps from the Boreas dataset \cite{Burnett2023}, in which we simulate a lidar scan from real lidar maps by subsampling the pointcloud. 
In this evaluation, \SI{25}{\percent} of the lidar scan is corrupted and perturbed using our framework, as this value is the highest resilience recorded in the experiments presented in \autoref{subsec:map_certif}.
Then, we feed this corrupted pointcloud to a vanilla \ac{ICP} and compare the produced error with the theoretical one given by \eqref{eq:error_faults}.
Vanilla \ac{ICP} is equipped with a trimmed distance filter, set to the same value $d$ as in our framework.
The initial guess for \ac{ICP} is set to the ground truth.

\begin{figure}[t]
  \centering
  \includegraphics[width=\linewidth]{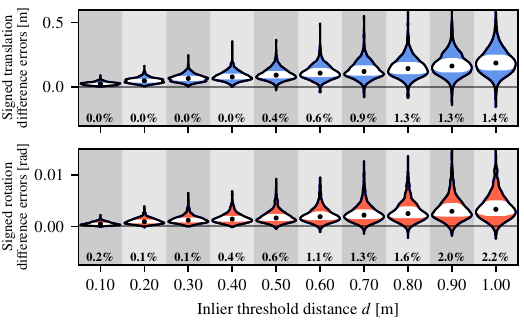}
  \caption{Signed difference of \ac{ICP} errors between our theoretical estimate and real \ac{ICP} for different inlier threshold distances.
  The violins depict the overall distribution of the difference.
   The medians are represented as black dots and the inner \SI{50}{\percent} of the data is depicted in shaded white.
	Bold numbers under the violins correspond to the associated ratio of false negative (\ie negative signed error).
    Our estimate typically yields larger errors compared to real \ac{ICP}, making it a reasonable approximation of an upper bound.
  }
  \label{fig:eval_icp}
  \vspace{-.5cm}
\end{figure}

\autoref{fig:eval_icp} depicts the signed difference between the theoretical errors predicted by our framework, and the real \ac{ICP} errors generated from the corrupted scans, for different inlier threshold distances $d$.
Overall, the difference in error exhibits a linear increase as the threshold distance $d$ increases. 
It can be seen that our framework yields higher errors (\ie over-estimates) compared to those obtained by the \ac{ICP} algorithm in a vast majority of time, with the ratio of false negative increasing along the outlier distance threshold.
In cases where our framework predicts a lower error than the real \ac{ICP} algorithm produces, the magnitude of error underestimation is on the order of centimeters and milliradians.
As shown in the next section, our framework provides an estimate that is tight enough to pinpoint safe and hazardous regions within the maps.
Future work will focus on providing a provable, tight upper bound.

For large inlier threshold distances, our framework overestimates the linear error up to $\SI{80}{\cm}$ in rare cases.
We theorize this discrepancy can be attributed to the fact that our framework assumes known data association and optimizes the perturbations accordingly.
As the inlier threshold distance increases, the corrupted points deviate further from their true associated points in the map.
Consequently, the likelihood of incorrect data associations also rises.
In some instances where wrong data association occurs, it unexpectedly favours the \ac{ICP} algorithm, as the perturbations applied were not optimized for that particular association.
As a result, an overall reduction in the pose error may occur.
On the other hand, the difference in rotational errors remains small.
Owing to the nature of the environment, the considered scans contain discernible features far away from the robot.
This makes it harder to corrupt the rotational component as compared to the translational one.

As such, assuming a known data association yields a reasonably conservative estimation on the error of \ac{ICP}.
This assumption is sensible as long as the environment features clear planes, as incorrectly associated points with the same plane will not affect the error terms.
However, in the case of noisy normal estimates or unstructured objects, an incorrect data association can match points associated with different planes and thus returns a different error term.
Such behavior is seen when our framework corrupts sectors containing bushes on the side of the road, or any other highly unstructured obstacles.
These cases are responsible for the long tails of the distributions seen in \autoref{fig:eval_icp}.
\subsection{Map Certification}
\label{subsec:map_certif}
To demonstrate the applicability of our framework for a given lidar map, we analyze real-world maps of urban, unstructured, and underground environments.
We certify the \texttt{Glen Shields} trajectory from the Boreas dataset \cite{Burnett2023}, trajectory \texttt{A} of the {Montmorency Forest Wintertime} dataset \cite{Baril2022}, and the map from the DARPA Subterranean challenge finals \cite{ebadi2022present}.
We sample poses along the trajectory and simulate live scans by subsampling the map for each pose. 
The scan is then used to build the matrices in \eqref{eq:ICP_ptpl}.
The outlier filter distance is $d=\SI{30}{\cm}$ and the lidar measurement noise $\bw$ has a standard deviation of $\sigma=\SI{10}{\cm}$.
These parameters are representative of realistic values used in autonomous driving.
The lidar field of view is split into \SI{30}{} sectors of approximately \SI{0.1}{\radian}.
An estimate is deemed unsafe if its probability to be outside the safe zone is above \SI{1}{\percent}.

For each pose, the degree of resilience $R$ is computed.
First, we provide a qualitative analysis of the Boreas and Forest dataset for a safe zone consisting of a \SI{20}{\cm} radius \mbox{L-$\infty$} ball in the $x$ and $y$ direction.
Then, a quantitative summary of the resilience over all the presented datasets for different safety requirements is presented.

\begin{figure}[htbp]
  \vspace{-1em}
  \centering
  \includegraphics[width=\linewidth]{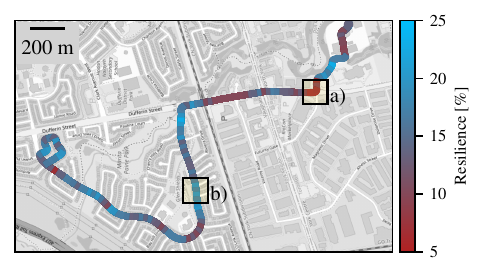}
  \vspace{-2.2em}%
  \caption{\texttt{Glen Shields} trajectory of the {Boreas} dataset, colored by its resilience to corruption $R$.
    The vehicle starts on the top right of the map, before driving on a large main road.
    It then leaves the road for smaller streets with better features.
	Highlighted places a) and b) correspond to the left and right pictures of \autoref{fig:intro}.
	}
  \label{fig:map_boreas}
  \vspace{-.5em}
\end{figure}

\autoref{fig:map_boreas} provides an aerial view of the \texttt{Glen Shields} trajectory along a detailed urban map of the environment.
The trajectory of the robot is colored by the degree of resilience at each submap.
The vehicle starts in a parking lot before turning onto a main street (\autoref{fig:map_boreas}.a). 
Then, the vehicle drives on smaller streets with better structural definition (\autoref{fig:map_boreas}.b).
Overall, the map is robust to an average corruption of around \SI{15}{\percent} of the pointcloud. %
When turning onto the main road, the vehicle has to cross a large crossroad with almost no structure present in the pointcloud.
As such, a small number of faults (\SI{6.5}{\percent}) are enough to push the estimate outside the safe bounds.
While on the main street, few structures are close enough to help constrain ICP in the longitudinal direction, resulting in resilience of around \SI{12}{\percent}.
Once the vehicle is on smaller streets with a lot of structural definition, \ac{ICP} becomes resilient to a greater amount of corruption.
Even if \SI{20}{\percent} of the scan is corrupted, there is still enough structure left untouched in the pointcloud to constrain \ac{ICP}.

\begin{figure}[t]
  \vspace{-.5em}
  \centering
  \includegraphics[width=\linewidth]{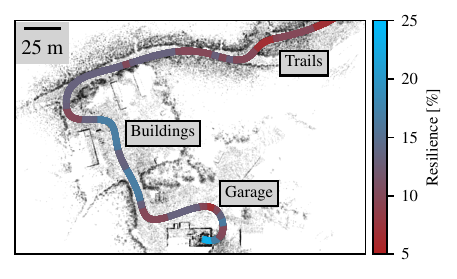}
  \vspace{-2.2em}%
  \caption{Path \texttt{A} of the {Montmorency Forest Wintertime} dataset, colored by its resilience to corruption $R$.
  The robot starts in a garage at the bottom of the map, proceeds to drive by several buildings, and finally enters a narrow ski trail surrounded by tall pine trees.
  }
  \label{fig:map_norlab}
  \vspace{-1em}
\end{figure}

Additionally, we analyze a trajectory of the {Montmorency Forest Wintertime} dataset, which is collected in subarctic environments in Northern Quebec, Canada, and consists of both semi-structured surroundings and unstructured trails in wintertime.
\autoref{fig:map_norlab} depicts the resulting resilience analysis on run \texttt{A} of the dataset.

The robot starts at the bottom of the map inside a garage.
It then drives next to some buildings before accessing a ski trail surrounded by tall pine trees.
Overall, the resilience to faults is on the same order as in urban environments, as long as the robot remains close to structures, being around \SI{15}{\percent}.
However, some key events can be noted that decrease the resilience.
As the robot leaves the garage, the only structured obstacle is one wall of the building, as all other surrounding obstacles are snowbanks.
Since \ac{ICP} relies heavily on one locally concentrated obstacle, a fault in this part of the map leads to a large error in the pose estimate, dropping the resilience to \SI{10}{\percent}.
As noted in \cite{Baril2022}, such faulted measurements were in fact observed during the recording of the dataset:
a truck parked next to the wall resulted in the measurements being offset by a small value, yielding the same type of corruption as the one theorized in this paper.
Harsh weather can also create similar effects, such as in the case of snow accumulating next to a building after a snowstorm overnight.
Once the robot drives toward the buildings, the resilience increases to \SI{15}{\percent}, mimicking an urban-like environment and thus its resilience.
The \ac{ICP} algorithm can indeed rely on a diverse source of information to estimate the pose.
Finally, once the robot reaches the trails, the resilience tends to drop down below \SI{10}{\percent} as it loses sight of the buildings and no other structured obstacles can help the localization process.

Finally, \autoref{tab:results} provides the overall statistics of each of the three environments discussed in this paper, for different safety requirements.
The first requirement of \SI{20}{\cm} along both the $x$ and $y$ components is the same as the one used in the qualitative experiments.
In general, the urban (Boreas) environment is safer, and the DARPA environment is the most prone to hazardous localization.
Indeed, the DARPA environment consists of long tunnels with few landmarks, making it extra hard for \ac{ICP} to generate a safe estimate in the presence of faults.
In the second safety requirements, the longitudinal component $x$ has been increased to \SI{50}{\cm}, emphasizing that shifting along the road direction is less dangerous.
With this relaxed requirement, both urban and DARPA environments see a clear increase in resilience, whereas the Forest environment does not have any apparent improvement, likely due to the lack of tunnel-like sections.
Finally, the third scenario requires \SI{20}{\cm} on the $x$ and $y$ components as well as a strong requirement on the yaw angle of \SI{0.05}{\radian}.
While a slight decrease in resilience can be seen in all three environments, both Boreas and DARPA see their minimal resilience drop to zero, meaning that there are some sections of the map where little to no corruption would make the robot have a hazardous estimate.
For the {Boreas} dataset, these sections correspond to places where the vehicle in on the large main road and little features are available.
The DARPA dataset is also more sensitive to such requirements because of the smaller scale of the environment, making it more challenging to constrain the rotational components.

\begin{table}[t]
	\centering
	\caption{Overall resilience statistics of the urban ({Boreas}), forest ({Montmorency Forest Wintertime}), and subterranean (DARPA) environments for different safety bounds $r_x,r_y,r_\theta$ on the longitudinal, lateral directions and yaw angle. Resilience values are reported in percents [\SI{}{\percent}].}
	\setlength\tabcolsep{2.5pt}
	\begin{tabularx}{\linewidth}{Xrrrrrrrrr}%
		\toprule
		&\multicolumn{3}{c}{$\begin{aligned}r_x&=\SI{20}{\cm}\\[-.3em]r_y&=\SI{20}{\cm}\end{aligned}$} 
		&\multicolumn{3}{c}{$\begin{aligned}r_x&=\SI{50}{\cm}\\[-.3em]r_y&=\SI{20}{\cm}\end{aligned}$} 
		&\multicolumn{3}{c}{$\begin{aligned}r_x&=\SI{20}{\cm}\\[-.3em]r_y&=\SI{20}{\cm}\\[-.3em]r_\theta&=\SI{0.05}{\radian}\end{aligned}$}\\
		\cmidrule(lr){2-4}\cmidrule(lr){5-7}\cmidrule(lr){8-10}
		&mean&std&min/max&mean&std&min/max&mean&std&min/max \\
		\midrule
		Urban & 14.9 & 4.0 & 6.7/23.3 & 17.3 & 4.2 & 6.7/26.7 & 13.1 & 6.9 & 0.0/23.3\\
		Forest & 12.1 & 3.0 & 6.7/23.3 & 12.6 & 2.8 & 6.7/23.3 & 12.0 & 3.1 & 6.7/23.3\\
		DARPA  & 8.4  & 3.7 & 3.3/16.7 & 10.4 & 4.1 & 3.3/20.0 & 6.0  & 4.4 & 0.0/13.3\\
		\bottomrule
	\end{tabularx}
	\label{tab:results}
	\vspace{-1em}
\end{table}

In conclusion, our resilience analysis is able to highlight dangerous locations where a small number of faults, coming from occlusion, noise, or changes in the environment, could drastically hinder the localization process and lead to hazardous behaviors.
These situations can either come from a clear lack of good features in the environment, or result from \ac{ICP} relying too much on a single environmental feature.

\section{Conclusion}
  In this paper, we present a novel way to analyze the resilience of the \ac{ICP} algorithm.
Resilience is defined as the maximum amount of faults that can be injected into the measurements before the localization estimate is dangerous for the robot.
We model faults as the most severe modifications to the measurements that can go undetected by an outlier filter.
Through this framework, we verify the quality of maps in both structured and unstructured environments, and demonstrate that environments lacking distinct, evenly distributed structures are more susceptible to inaccurate estimates in the event of measurement corruption.

Future research will focus on addressing the assumptions made in this paper.
We will account for the iterative nature of the \ac{ICP} algorithm and explore a broader range of robust cost functions.
Also, this paper assumes that the robot has a good initial guess. 
However, hazardous behaviors can also occur during a sequence of bad \ac{ICP} solutions, each driving the robot farther from the ground truth and feeding worse and worse initial guesses to the next iteration.
As such, taking into consideration a possibly wrong initial guess will also help certify against a broader range of failures.
Finally, future work will also explore the use of this framework in landmark-based localization approaches.

\section*{ACKNOWLEDGMENT}
We would like to thank the Natural Sciences and Engineering Research Council of Canada (NSERC) and the Ontario Research Fund: Research Excellence (ORF-RE) program for supporting this work.
We also thank the Northern Robotics Laboratory (Norlab) for their help with the datasets.

\renewcommand*{\bibfont}{\footnotesize}
\printbibliography

\end{document}